\documentclass{Interspeech2024}

\interspeechcameraready

\title{Developing an End-to-End Framework for Predicting the Social Communication Severity Scores of Children with Autism Spectrum Disorder}

\name[affiliation={1}]{Jihyun}{Mun}
\name[affiliation={2}]{Sunhee}{Kim}
\name[affiliation={1}]{Minhwa}{Chung}

\address{
  $^1$Department of Linguistics, Seoul National University, Republic of Korea\\
  $^2$Department of French Language Education, Seoul National University, Republic of Korea}
\email{jhhh\_1202@snu.ac.kr, sunhkim@snu.ac.kr, mchung@snu.ac.kr}


\usepackage{booktabs}
\usepackage{multirow}
\usepackage{xcolor}

\begin{document}

\maketitle

\begin{abstract}
Autism Spectrum Disorder (ASD) is a lifelong condition that significantly influencing an individual's communication abilities and their social interactions. Early diagnosis and intervention are critical due to the profound impact of ASD's characteristic behaviors on foundational developmental stages. However, limitations of standardized diagnostic tools necessitate the development of objective and precise diagnostic methodologies. This paper proposes an end-to-end framework for automatically predicting the social communication severity of children with ASD from raw speech data. This framework incorporates an automatic speech recognition model, fine-tuned with speech data from children with ASD, followed by the application of fine-tuned pre-trained language models to generate a final prediction score. Achieving a Pearson Correlation Coefficient of 0.6566 with human-rated scores, the proposed method showcases its potential as an accessible and objective tool for the assessment of ASD.
\end{abstract}
\section{Introduction}

Autism Spectrum Disorder (ASD) is defined as a lifelong condition that significantly affects an individual's communication abilities and their interaction within society \cite{al2015descriptive}. 
Children with ASD experience social deficits, communication difficulties, and atypical behavior patterns, including impaired socio-communicative interactions and a limited range of interests and activities \cite{al2015descriptive, nicholas2008prevalence}.

Early diagnosis and intervention are critical due to the profound impact of ASD's symptomatic behaviors on foundational developmental processes. Early intervention is particularly pivotal for social development, as initial social capabilities and deficits inform intervention outcomes and treatment strategies \cite{lord2006early, zachor2010treatment}.
In clinical environments, standardized diagnostic tools like the Autism Diagnostic Observation Schedule, 2nd edition (ADOS-2), are employed \cite{lord_autism_2012}. 
However, the use of standardized tools for evaluating children presents numerous challenges, including expertise scarcity leading to delayed or overlooked diagnoses \cite{li2022automatic}, potential bias from subjective interpretations by caregivers or evaluators \cite{frigaux2019adiados}, and the extended duration of the evaluation process, which can burden both children and their caregivers and may reduce the children's concentration.
Consequently, there is a pressing need for developing objective and precise methodologies which diagnose and predict severity for early diagnosis and intervention of ASD \cite{wang2022mage, hadoush2019automated}.

Recent advancements in automated methods for predicting ASD severity incorporate a range of technologies, including MRI \cite{moradi2017predicting, uddin2013salience, kwon2022sparse}, fMRI \cite{wang2021metamodel}, EEG signals \cite{hadoush2019automated, zhang2022predicting}, and genetic and environmental factors \cite{che2019approach}. 
Despite their efficacy, these methods often require specialized equipment and expertise, presenting barriers to widespread adoption \cite{li2022automatic}. In contrast, speech data offers a more accessible and less intrusive alternative \cite{clemente2008recording}, providing a viable option for diagnosing and assessing the severity of ASD. 
Studies have concentrated on the pragmatic aspects of language, including the appropriate use of language across various social contexts, particularly in children with ASD in comparison to their typically developing (TD) peers \cite{philofsky2007pragmatic, volden2009brief, vogindroukas2022pragmatics}.
They underscored that children with ASD frequently exhibit atypical language behaviors in social contexts, thereby emphasizing the complex relationship between linguistic and social challenges.
The utilization of speech data not only circumvents the limitations associated with other diagnostic materials but also leverages the unique linguistic characteristics of children with ASD.
This underscores the potential of linguistic materials for the automated diagnosis and severity prediction of ASD \cite{cho2019automatic, ashwini2023spasht, chen2020learning}, offering a promising direction for enhancing accessibility and reducing the reliance on extensive resources and specialized knowledge.

Machine learning techniques have been applied to identify ASD based on linguistic indicators \cite{cho2019automatic, ashwini2023spasht}, with traditional methods requiring meticulous feature selection, a process that is time-intensive and highly specialized \cite{verdonck2021special}. Deep learning approaches offer an alternative by deriving more abstract representations \cite{bengio2013representation}, such as using lexical embeddings from a fine-tuned BERT model for ASD diagnosis \cite{chen2020learning}. However, deep learning models necessitate large datasets, which poses a challenge for ASD research due to the typically small available datasets. Pre-trained language models (PLMs), fine-tuned on specific tasks, leverage extensive pre-training corpora to mitigate this issue \cite{devlin2018bert}.

A notable concern when applying PLMs to classification tasks is the potential misalignment between the objectives during pre-training and fine-tuning \cite{wang2023exploiting}. 
The integration of natural language prompts in fine-tuning PLMs, a technique known as prompt tuning, aligns the model’s objectives with those of the pre-training phase, thereby enhancing performance on specific tasks in the context of limited data \cite{wang2023exploiting, chang2023speechprompt}. 

Building on recent methodological advancements and leveraging the distinctive benefits of prompt tuning in contexts with limited data, this paper proposes an end-to-end (E2E) framework that incorporates a prompt tuning methodology for predicting the severity of social communication in children with ASD.
The deployment of prompt tuning methodologies necessitates the transcription of audio recordings. However, manual transcription presents several challenges, including high costs, limited availability, and issues with scalability.
To overcome these challenges, we integrate an Automatic Speech Recognition (ASR) model into our framework, enabling the derivation of final prediction scores directly from raw speech data.
The comprehensive framework utilizes an ASR model, specifically fine-tuned with speech data from children with ASD, followed by the application of fine-tuned PLMs and an ensemble method to generate a final prediction score.


The remainder of the paper is organized as follows: Section 2 details the methodologies employed, Section 3 outlines the experimental setup, Section 4 presents the results, Section 5 discusses the findings, and Section 6 concludes the study.
\section{Methods}
\begin{figure}[t]
  \centering
  \includegraphics[width=\linewidth]{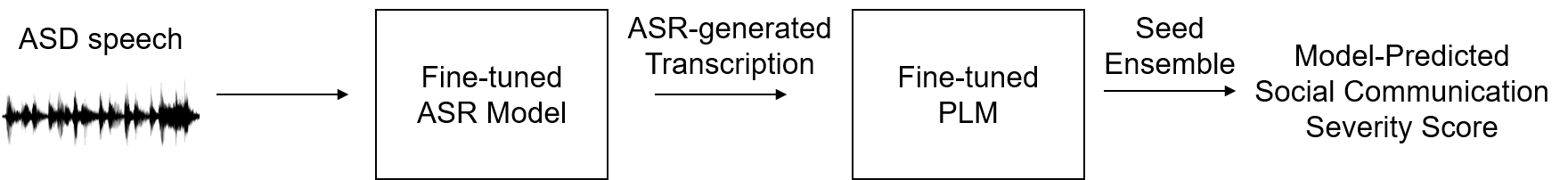}
  \caption{Proposed E2E framework for automatically predicting the social communication severity scores of children with ASD}
  \label{fig:framework}
\end{figure}

This study introduces an E2E framework that incorporates fine-tuned ASR models, fine-tuned PLMs, and a seed ensemble method for predicting the social communication severity scores in children with ASD, as depicted in Figure~\ref{fig:framework}.

\subsection{Automatic Speech Recognition Model}
We selected two pre-trained multilingual ASR models for this purpose: wav2vec2-xls-r-300m \cite{babu2021xls} and whisper-large-v2 \cite{radford2022whisper}. To tailor these models to the nuances of speech from TD children and children with ASD, we fine-tune each model using speech data specific to these groups.

\subsection{Fine-tuning Pre-trained Language Models}
The study further involves fine-tuning three PLMs—KR-BERT \cite{lee2020krbert}, KLUE/roberta-base \cite{park2021klue}, and KR-ELECTRA-Discriminator \cite{kr-electra}-employing three distinct approaches: traditional fine-tuning, manual prompting, and p-tuning. These models were chosen for prompt-based fine-tuning due to their demonstrated effectiveness in text classification tasks, as evidenced by prior research \cite{wang2023exploiting, wang2023medical, yao2022prompt}. Training incorporates ten different initialization seeds to increase robustness and mitigate the effects of random initialization.

\subsubsection{Traditional Fine-tuning}
Fine-tuning adapts a model pre-trained on a vast dataset to a smaller, task-specific dataset \cite{howard2018universal}, effectively leveraging the extensive knowledge acquired during pre-training \cite{devlin2018bert} for specific downstream tasks. In this process, a regression head is attached to the model. The [CLS] token, representing the input sequence comprehensively, facilitates the prediction of a continuous severity score.

\subsubsection{Manual Prompting}
Manual prompting involves crafting specific input prompts to direct the behavior of transformer models towards generating desired outputs. By designing appropriate prompts, it's possible to utilize the extensive knowledge embedded in these models for performing specific tasks without additional task-specific training. 
In this approach, a regression head initializes the models, and a template guides the model to focus on predicting the social communication severity score from the input text.

\subsubsection{Automated Prompting: P-Tuning}
P-tuning, introduced by \cite{liu2023gpt}, advances beyond manual prompting by parameterizing prompts and optimizing them alongside the model's parameters during fine-tuning, allowing the model to autonomously identify the most effective prompts for a task. 
For this study, the p-tuning approach is implemented using the PEFT library \cite{peft}. Models are initialized with a regression head, and virtual tokens are incorporated and tuned specifically for the task, optimizing the models' predictions.

\subsection{Seed Ensemble for Robust Prediction}
To mitigate the variability introduced by the randomness in model initialization and to improve the overall performance, a seed ensemble technique is employed. For each PLM, we aggregate the predictions from the ten individually fine-tuned models (one per seed) to formulate a singular and more accurate prediction.

\section{Experiments}

\subsection{Data Preparation and Dataset Description}
The speech samples were collected during linguistic assessment sessions conducted by certified speech-language pathologists (SLPs). The specifics of the data collection, transcription, and evaluation processes have been detailed in \cite{lee2024speech}. 
This study utilized speech data from 168 children diagnosed with ASD and 40 TD children. These participants were integral for fine-tuning the ASR models. Specifically, the ASD cohort included 103 children whose social communication severity was evaluated by three certified SLPs. The average of the three SLPs' evaluations served as the severity score for the ASD children, while TD children were assigned a baseline score of zero. The datasets for evaluated ASD and TD children were employed for fine-tuning the PLMs.
The overall dataset is described in Table~\ref{data}. To ensure no overlap and maintain the integrity of the evaluation process, children included in the test set for PLM fine-tuning were excluded from the training dataset of the ASR model.

\begin{table}[]
\centering
\caption{ASR and PLM fine-tuning speakers}
\label{data}
\resizebox{\columnwidth}{!}{%
\begin{tabular}{@{}ccccc@{}}
\toprule
\multirow{2}{*}{} & \multicolumn{2}{c}{Train} & \multicolumn{2}{c}{Test} \\ \cmidrule(l){2-5} 
                  & ASD          & TD         & ASD         & TD         \\ \midrule
\begin{tabular}[c]{@{}c@{}}ASR\\ tuning\\ data\end{tabular} &
  \begin{tabular}[c]{@{}c@{}}152\\ (11h 15m 10s)\end{tabular} &
  \begin{tabular}[c]{@{}c@{}}32\\ (2h 54m 51s)\end{tabular} &
  \begin{tabular}[c]{@{}c@{}}16\\ (2h 14m 53s)\end{tabular} &
  \begin{tabular}[c]{@{}c@{}}8\\ (35m 48s)\end{tabular} \\ \midrule
\begin{tabular}[c]{@{}c@{}}PLM\\ tuning\\ data\end{tabular} &
  \begin{tabular}[c]{@{}c@{}}87 \\ (73 M, 13 F, 1 U)\\ (mean 9;1, std 5;8)\end{tabular} &
  \begin{tabular}[c]{@{}c@{}}32\\ (13 M, 11 F, 8 U)\\ (mean 5;11, std 2;10)\end{tabular} &
  \begin{tabular}[c]{@{}c@{}}16\\ (13 M, 3 F)\\ (mean 12;11, std 7;3)\end{tabular} &
  \begin{tabular}[c]{@{}c@{}}8\\ (1 M, 3 F, 4 U)\\ (mean 4;1, std 2)\end{tabular} \\ \bottomrule
  \multicolumn{5}{r}{\footnotesize M: male (boy), F: female (girl), U: unreported} \\
  \multicolumn{5}{r}{\footnotesize mean and std refer to the mean and standard deviation of chronical age of speakers} \\
\end{tabular}%
}
\end{table}




\subsection{Fine-Tuning ASR Models}
The ASR models, specifically wav2vec2 and whisper, are fine-tuned using Fairseq and Hugging Face's Transformers, respectively. The Adam optimizer is utilized in both cases, with initial learning rates set to 3e-4 for wav2vec2 and 1e-5 for whisper. Given that Korean is a syllable-timed language, the performance of the fine-tuned models is evaluated using the syllable error rate (SER), achieving rates of 26.21\% and 19.57\%, respectively, after fine-tuning.

\subsection{Fine-Tuning PLMs}
For each tuning method, training spanned 40 epochs, utilizing a learning rate of 1e-5, a batch size of 8, and the AdamW optimizer. The mean squared error is employed as the objective loss function. In manual prompting, the template "[text] the social communication severity score of the speaker is [MASK]" is used, with "[text]" replaced by actual dataset text and "[MASK]" serving as a placeholder. In p-tuning, experiments are conducted with 5, 10, 15, and 20 virtual tokens, setting the encoder's hidden size to 128. Differential learning rates are applied: 1e-5 for both the base models and the prompt encoder, and 1e-3 for the regression head.

\subsection{Evaluation Metrics}
The evaluation strategy includes two settings:
\begin{enumerate}
\item \textit{Full-set setting}, where all available training data is used, reserving 20\% for validation.
\item \textit{Low-resource setting}, where only 20\% of the full training data is accessible, following the methodology outlined by \cite{yao2022prompt}.
\end{enumerate}
The evaluation metric employed is the Pearson Correlation Coefficient (PCC), which measures the relationship between the model's predicted output and the scores labeled by humans. To mitigate the effects of random initialization, each system's evaluation is executed ten times, each with a different random seed from PyTorch's random initialization setting. The final prediction is determined using the seed ensemble method.

\section{Results}
The study evaluates the effectiveness of the proposed framework, which integrates various ASR models, transcription types, PLMs, and tuning methods in predicting social communication severity in children with ASD across full-set and low-resource settings.
The comprehensive results of our experiments are shown in Table~\ref{results}.

As expected, human transcriptions consistently outperform ASR transcriptions. 
However, certain combinations of PLMs and tuning methods, specifically klue/roberta-base with p-tuning, reveal instances where ASR transcriptions surpass human transcriptions.
In low-resource settings, the performance gap between human and ASR transcriptions diminishes, highlighting the potential of ASR transcriptions in scenarios of limited data availability. Remarkably, wav2vec2 transcription outperforms human transcription in specific cases when klue/roberta-base model is p-tuned, indicating a strong correlation with human-labeled scores (e.g., PCC of 0.6566 compared to 0.6216 with 20 virtual tokens).
When comparing two ASR models, wav2vec2 transcriptions generally exhibit better performance than those from the whisper model, despite a higher syllable error rate. 

The results demonstrate that the choice of PLM and the tuning method significantly affects the performance in predicting the severity score of social communication.
In scenarios involving both ASR and human transcriptions within the full-set setting, fine-tuning and manual prompting tend to outperform p-tuning for the KR-BERT and KR-ELECTRA-Discriminator models. However, p-tuning shows superior performance with the klue/roberta-base model. 
This trend continues in the low-resource setting, where p-tuning enhances performance with human transcriptions for the KR-ELECTRA-Discriminator model.

Additionally, performance varies significantly based on the number of virtual tokens utilized in p-tuning. 
For example, with the KR-BERT model using ASR transcriptions, the PCC values range from negative to positive, indicating a shift from a negative to a moderate correlation with human-labeled scores.
Similarly, with the KR-BERT model using human transcriptions in a low-resource setting, the correlation varies significantly from weak to moderate.

\begin{table*}[t]
\caption{Pearson correlation coefficient with human-labeled scores}
\label{results}
\begin{tabular}{@{}ccccccccc@{}}
\toprule
\multicolumn{3}{c}{\multirow{3}{*}{}}                                                                                                            & \multicolumn{3}{c}{\textbf{\begin{tabular}[c]{@{}c@{}}Full-set setting\end{tabular}}}                                                                                            & \multicolumn{3}{c}{\textbf{\begin{tabular}[c]{@{}c@{}}Low-resource setting\end{tabular}}}                                                                                        \\ \cmidrule(l){4-9} 
\multicolumn{3}{c}{}                                                                                                                             & \multicolumn{2}{c}{\textbf{\begin{tabular}[c]{@{}c@{}}ASR\\ transcription\end{tabular}}} & \multirow{2}{*}{\textbf{\begin{tabular}[c]{@{}c@{}}Human\\ transcription\end{tabular}}} & \multicolumn{2}{c}{\textbf{\begin{tabular}[c]{@{}c@{}}ASR\\ transcription\end{tabular}}} & \multirow{2}{*}{\textbf{\begin{tabular}[c]{@{}c@{}}Human\\ transcription\end{tabular}}} \\ \cmidrule(lr){4-5} \cmidrule(lr){7-8}
\multicolumn{3}{c}{}                                                                                                                             & \textbf{Wav2vec2}                           & \textbf{Whisper}                           &                                                                                         & \textbf{Wav2vec2}                           & \textbf{Whisper}                           &                                                                                         \\ \midrule
\multirow{6}{*}{\textbf{KR-BERT}}                                                             & \multicolumn{2}{c}{\textbf{Fine-tuning}}         & 0.2791                                      & 0.1984                                     & 0.5516**                                                                                & 0.4471*                                     & 0.2253                                     & 0.5817**                                                                                \\ \cmidrule(l){2-9} 
                                                                                              & \multicolumn{2}{c}{\textbf{Manual}}              & 0.3637                                      & 0.1624                                     & 0.4701*                                                                                 & 0.4204*                                     & 0.0869                                     & 0.5032**                                                                                \\ \cmidrule(l){2-9} 
                                                                                              & \multirow{4}{*}{\textbf{P-tuning}} & \textbf{5}  & -0.1992                                     & -0.1576                                    & 0.4483*                                                                                 & 0.1808                                      & -0.0346                                    & 0.1119                                                                                  \\ \cmidrule(l){3-9} 
                                                                                              &                                    & \textbf{10} & 0.3861                                      & 0.2129                                     & 0.4511*                                                                                 & 0.3815                                      & 0.2841                                     & 0.5595**                                                                                \\ \cmidrule(l){3-9} 
                                                                                              &                                    & \textbf{15} & 0.3367                                      & 0.1077                                     & 0.3139                                                                                  & 0.1491                                      & -0.0602                                    & 0.3409                                                                                  \\ \cmidrule(l){3-9} 
                                                                                              &                                    & \textbf{20} & -0.0047                                     & -0.0410                                    & 0.4663*                                                                                 & -0.2881                                     & -0.1187                                    & 0.0050                                                                                  \\ \midrule
\multirow{6}{*}{\textbf{klue/RoBERTa-base}}                                                   & \multicolumn{2}{c}{\textbf{Fine-tuning}}         & 0.3880                                      & 0.2070                                     & 0.4322*                                                                                 & 0.3806                                      & 0.2271                                     & 0.3972                                                                                  \\ \cmidrule(l){2-9} 
                                                                                              & \multicolumn{2}{c}{\textbf{Manual}}              & 0.0761                                      & 0.0846                                     & 0.2859                                                                                  & 0.3515                                      & 0.1445                                     & 0.4367*                                                                                 \\ \cmidrule(l){2-9} 
                                                                                              & \multirow{4}{*}{\textbf{P-tuning}} & \textbf{5}  & 0.5587**                                    & 0.4980*                                    & 0.4207*                                                                                 & 0.6117**                                    & 0.5633**                                   & 0.5731**                                                                                \\ \cmidrule(l){3-9} 
                                                                                              &                                    & \textbf{10} & 0.5431**                                    & 0.5109*                                    & 0.5181**                                                                                & 0.6333***                                   & 0.6183**                                   & 0.6217**                                                                                \\ \cmidrule(l){3-9} 
                                                                                              &                                    & \textbf{15} & 0.5343**                                    & 0.4331*                                    & 0.4812*                                                                                 & 0.6163**                                    & 0.6166**                                   & 0.6230**                                                                                \\ \cmidrule(l){3-9} 
                                                                                              &                                    & \textbf{20} & 0.5852**                                    & 0.5330**                                   & 0.5472**                                                                                & 0.6566**                                    & 0.5854**                                   & 0.6216**                                                                                \\ \midrule
\multirow{6}{*}{\textbf{\begin{tabular}[c]{@{}c@{}}KR-ELECTRA\\ -Discriminator\end{tabular}}} & \multicolumn{2}{c}{\textbf{Fine-tuning}}         & 0.4649*                                     & 0.4315*                                    & 0.9019***                                                                               & 0.3425                                      & 0.1735                                     & 0.6454***                                                                               \\ \cmidrule(l){2-9} 
                                                                                              & \multicolumn{2}{c}{\textbf{Manual}}              & 0.4452*                                     & 0.2109                                     & 0.7645***                                                                               & 0.4207*                                     & 0.1556                                     & 0.6925***                                                                               \\ \cmidrule(l){2-9} 
                                                                                              & \multirow{4}{*}{\textbf{P-tuning}} & \textbf{5}  & 0.0750                                      & 0.0605                                     & 0.6546***                                                                               & 0.1652                                      & 0.0564                                     & 0.7138***                                                                               \\ \cmidrule(l){3-9} 
                                                                                              &                                    & \textbf{10} & -0.1221                                     & 0.0485                                     & 0.6509***                                                                               & 0.2095                                      & 0.1134                                     & 0.7117***                                                                               \\ \cmidrule(l){3-9} 
                                                                                              &                                    & \textbf{15} & 0.0335                                      & 0.0491                                     & 0.7164***                                                                               & 0.1552                                      & 0.0948                                     & 0.7273***                                                                               \\ \cmidrule(l){3-9} 
                                                                                              &                                    & \textbf{20} & -0.2002                                     & -0.0258                                    & 0.5335**                                                                                & 0.3324                                      & 0.2717                                     & 0.7654***                                                                               \\ \bottomrule

\multicolumn{9}{r}{\scriptsize \textit{*: p$<$0.05, **: p$<$0.01, ***: p$<$0.001}}
\end{tabular}
\end{table*}

\section{Discussion}

The results highlight a complex relationship between transcription types, PLM selection, tuning methods, and data availability in the automated assessment of ASD severity.

The diminishing performance disparity between human and ASR transcriptions in low-resource settings underscores the proposed method's potential in enhancing the accessibility and scalability of ASD severity assessment. This trend suggests that ASR technology may serve as a feasible alternative to human transcription in situations where resources are limited.


The generally better performance of the wav2vec2 model over the whisper model, despite the latter's lower error rate, indicate that there are aspects of speech relevant to ASD severity that are captured by wav2vec2 but ignored by whisper due to its disfluency removal.
It is known that children with ASD display various types of speech disfluencies, such as sound and syllable repetitions, interjections, within-word breaks, and final sound prolongations \cite{plexico2010disfluency}.
The whisper model's tendency to eliminate speech disfluencies, including filler words, hesitations, and repetitions \cite{lintoai2023whispertimestamped}, contrasts with the wav2vec2 model's capability to detect disfluencies or stuttering.
Therefore, accurately capturing the characteristics of ASD speech, including speech disfluencies, necessitates the selection of an appropriate ASR model that retains these critical speech features. This consideration is pivotal in developing effective diagnostic tools and interventions for ASD, highlighting the importance of choosing an ASR model that aligns with the nuanced requirements of ASD speech.

The varied performance across PLMs under different tuning methods highlights the necessity of meticulous consideration for each PLM-tuning combination. 
The klue/roberta-base model's effective response to p-tuning, across both transcription types, suggests its potential as a powerful tool in optimizing PLMs, particularly in data-constrained environments. 
Additionally, the number of virtual token significantly influences performance differences. Although the number of prompt tokens greatly impacts few-shot performance, a larger number of prompt tokens is not always better; it depends on the amount of training data \cite{liu2023gpt}. In practice, we should determine the optimal number of prompt tokens through model selection, highlighting the need for careful consideration of tuning settings.
\section{Conclusion}

This study proposes an E2E framework, incorporating fine-tuned ASR models and PLMs, for automatically predicting social communication severity in children with ASD.
Demonstrating a PCC of 0.6566, the experimental results affirm the framework's utility, especially in data-limited situations.

Key contributions of this paper include the introduction of an automated method for predicting the social communication severity score in children with ASD from raw speech data, the development of an E2E framework that eliminates the need for human transcription, and the validation of this framework's effectiveness in data-restricted settings. These achievements indicate the practical applicability of the framework in real-world ASD severity assessments, a field where acquiring large datasets is often challenging.

Despite these promising results, the framework faces interpretability challenges. In domains such as ASD diagnosis and assessment, the models' decision-making processes must be transparent to ensure trust and reliability in their practical application \cite{vellido2020importance}. Interpretability becomes even more crucial in extremely data-limited situations, where variability in results can be substantial. However, the highest-performing model employs P-tuning of the PLM, which utilizes virtual tokens as learnable parameters. These parameters are inherently non-interpretable and untrackable, obscuring the model's decision logic and further complicating the issue of interpretability.

Future research will explore instruction tuning methodologies that could provide "chain-of-thought" reasoning \cite{wei2022chain}, potentially enhancing the interpretability of model predictions.
The goal is to align high predictive accuracy with clear, understandable outputs, ensuring that the models not only predict with high precision but also provide interpretable and actionable insights for clinical use.

\section{Acknowledgements}
This work was supported by Institute of Information \& Communications Technology Planning \& Evaluation (IITP) grant funded by the Korea government(MSIT) [No.2022-0-00223, Development of digital therapeutics to improve communication ability of autism spectrum disorder patients].


\bibliographystyle{IEEEtran}
\bibliography{e2e}

\end{document}